# Accepted Manuscript

Communicating with sentences: A multi-word naming game model

Yang Lou, Guanrong Chen, Jianwei Hu

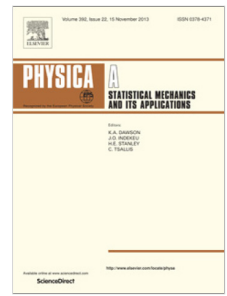

# Communicating with sentences: A multi-word naming game model

Yang Lou[1], Guanrong Chen[1*], and Jianwei Hu[2]

[1] Department of Electronic Engineering, City University of Hong Kong, Hong Kong SAR, China
[2] School of Electronic Engineering, Xidian University, Xi'an 710071, China

*Corresponding author: eegchen@cityu.edu.hk

**Abstract** – Naming game simulates the process of naming an object by a single word, in which a population of communicating agents can reach global consensus asymptotically through iteratively pair-wise conversations. We propose an extension of the single-word model to a multi-word naming game (MWNG), simulating the case of describing a complex object by a sentence (multiple words). Words are defined in categories, and then organized as sentences by combining them from different categories. We refer to a formatted combination of several words as a pattern. In such an MWNG, through a pair-wise conversation, it requires the hearer to achieve consensus with the speaker with respect to both every single word in the sentence as well as the sentence pattern, so as to guarantee the correct meaning of the saying; otherwise, they fail reaching consensus in the interaction. We validate the model in three typical topologies as the underlying communication network, and employ both conventional and man-designed patterns in performing the MWNG.

## 1. Introduction

Naming game (NG) is a simulation-based numerical study exploring the emergence of shared lexicon in a population of communicating agents about a same object which they observed [1–3]. The single object in NG can be an entity, idea, opinion, or a social or economic convention that can be described by a single word [4]. A population of self-organized agents is connected in a certain topology representing the relationships among them. The minimal NG is described as follows. Each time, one pair of connected speaker and hearer is randomly selected from the population. If the object is unknown to the speaker (who has no word to describe the object), he will invent a new name and then tell it to the hearer. When the object is known to the speaker, he will randomly pick a name from memory and utter it. When the hearer receives the name, he will search over his memory to see if he has the same name stored therein: if not, then he will store it into memory; but if yes, then they reach consensus, so both clear up all the names, while keeping the common one respectively. This pair-wise success is referred to as *local consensus* hereafter. Such a pair-wise transmitting and receiving (or teaching and learning) process will eventually lead the entire population of agents to reach consensus, referred to as *global consensus*, meaning all the agents agree to describe the object by the same name. The convergence property of NG is not only observed by numerical simulations, but also proved theoretically [5] and verified empirically [4]. This conventional NG will be called the single-word naming game (SWNG) below.

Previous studies on naming game focus on mainly two aspects: the agent dynamics [2,6–12] and the information dynamics [13–16]. The former concerns about the topological relationships of agents, the roles of speaker and hearer, and the communication model among the agents, as well as their characteristics. For example, the minimal NG is investigated on random-graph and scale-free networks in [6,7], and on small-world networks in [8,9]. The speaker-only naming game (SO-NG) and hearer-only naming game (HO-NG) are proposed in [10]. In SO-NG, only the speaker will update his memory, while in HO-NG only the hearer will do so. Later, the NG with multiple hearers (NGMH) was proposed in [11], which is an HO-NG model with an additional rule: only when all hearers reached consensus with the same speaker, they reach local consensus together. The NG in groups (NGG) [12] allows every agent in a selected group from the population to play the dual role as a speaker and also as a hearer. All the aforementioned models have demonstrated that the convergence speed is faster when the number of participating agents increases. The finite-memory NG (FMNG), proposed in [2], studies the situation where the memory sizes of the agents are limited.

The study of information dynamics includes: The case when communications are influenced by learning errors is studied in [13], and the memory loss case is studied in [16]. The combinational NG [14] decomposes single words as atomic names and combinational names. An atomic name is unique and independent of other atomic names, e.g., '*blue*' and '*sky*'. In contrast, a combinational name is a permuted combination of atomic names, e.g., '*sky-blue*' and '*blue-sky*'. The blending game discussed in [15] uses similar combinational forms, but with different communication rules, thus a blending name will be composed only if a speaker-uttered name is not agreed by the hearer.

Whether a name for the object is atomic or combinational is determined based on the coding (mapping) method from the set of atomic words into the set of combinational words [14]. An illustrative example is shown in Figure 1, where the integer coding could transform the atomic names to be combinational, and further, be decomposed into binary strings. This lacks a certain pattern to reflect the order and relationship between the components. Both atomic NG and combinational NG are SWNG, because the transmission information of each pair-wise interaction is a single word.

```
light 108 105 103 104 116
        l    i    g    h    t
right 114 105 103 104 116
        r    i    g    h    t
    l   (108)₁₀ = (01101100)₂
    r   (114)₁₀ = (01110010)₂
```

**Figure 1** An example of coding in naming game. Two atomic names, '*light*' and '*right*', can be decomposed into 5 independent letters respectively, where the integers are ASCII code for each letter. These two atomic names could be considered as '*l*' + '*ight*' and '*r*' + '*ight*', respectively. The difference between '*light*' and '*right*' are the initial words '*l*' and '*r*'. They can also be decomposed in other ways. When every letter in a name is coded in binary, the letters can be further decomposed into sequences of binary strings, e.g., '011', '0', and '1001'. As a result, '*l*' consists of two '011's and '0's in order, and '*r*' consists of one '011' followed by one '1001', with one '0' at the end.

In this paper, we propose a multi-word NG (MWNG) to study the scenario where the names of objects are described by neither atomic nor combinational words, but more complicatedly, by sentences. We first classify the names into several categories, and then define some patterns as sentential structures, so that a sentence can be composed following a certain pattern. Here, the pattern employed in MWNG is simple, since the focus is on the converging process, rather than on the grammatical analysis of languages as in [17,18]. Note that due to the duality caused by coding, we can consider a specific sentence as a single atomic name, thus the MWNG is degenerated to the SWNG, or minimal NG. In this sense, the MWNG is a natural extension of the conventional SWNG.

## 2. Methods

The distinguished attribute of the MWNG model is that the speaker utters several words simultaneously (a simple sentence) rather than a single word to describe an object (e.g., an opinion, an event, etc.), which is more realistic and common in human conversations.

The pattern of sentences in MWNG is kept simple, as long as it is able to show some organizing structure of the words. Actually, patterns could also be considered as simple grammar. A simple implementation of pattern could be formed by combination of words from different categories. Categorization of words directly affects the resolution of patterns. Words of the same category are of equivalent importance. The term resolution concerns with the precision and correctness of a sentence. For example, one may classify the words into as rough as three categories, '*noun*', '*verb*', '*adjective*', and define a pattern '*noun + verb + noun*'. This pattern can guide one to produce sentences like '*boys play football*'. Likewise a sentence '*football plays boys*' perfectly follows the same pattern, but practically meaningless. However, if the classification of word categories is as detailed as '*human*', '*human-action-verb*', '*sports-name*', etc., then as such a pattern like '*human + human-action-verb + sports-name*' is of high resolution to guide one to generate meaningful sentences in practice. In this case, '*football plays boys*' will be excluded from the same pattern of '*boys play football*'. Too few categories will probably lead to too many ambiguous or meaningless sentences, but too many categories are clearly inefficient.

This paper, therefore, studies five simple and conventional patterns of the English language, as well as other five sets of man-designed sentence patterns. Since the category classification of vocabulary is arbitrary [19], we assume that the category is associated with the word, so that an agent identifies the category of a word as soon as he has it (either invents it or receives it).

For simplicity, the following two assumptions are made: 1) a pattern of a sentence is a unique sequence of word categories; 2) a tag indicating the category is associated with a word, so that an agent can identify the category immediately. The natural assumptions make the model easy to understand. Each sentence has one and only one pattern without exception. As soon as an agent receives a sentence, he receives the pattern based on the category and the order of words. In real-life communications, however, a speaker would not offer a category tag along with each word he says, but some illustrative tags will be given in the communication. For example, when one tries to understand a received word '*chakalaka*' (an African food), he will naturally associate it with tags such as '*noun*', '*food*', '*exotic*', etc. Thus, a new word is categorized when it is received.

Figure 2 shows the flowchart of one time step in MWNG, where in the beginning a connected pair of speaker-hearer is picked. Either direct strategy [6,9] or reverse strategy [9] can be applied. In the direct strategy, a speaker is selected first, and then a hearer is randomly picked from the speaker's connected neighborhood. In the reverse strategy, it is opposite: a hearer is picked first, and then a speaker is picked from the hearer's connected neighborhood. Here, the direct strategy is used in MWNG, while the reverse strategy is discussed in SI.

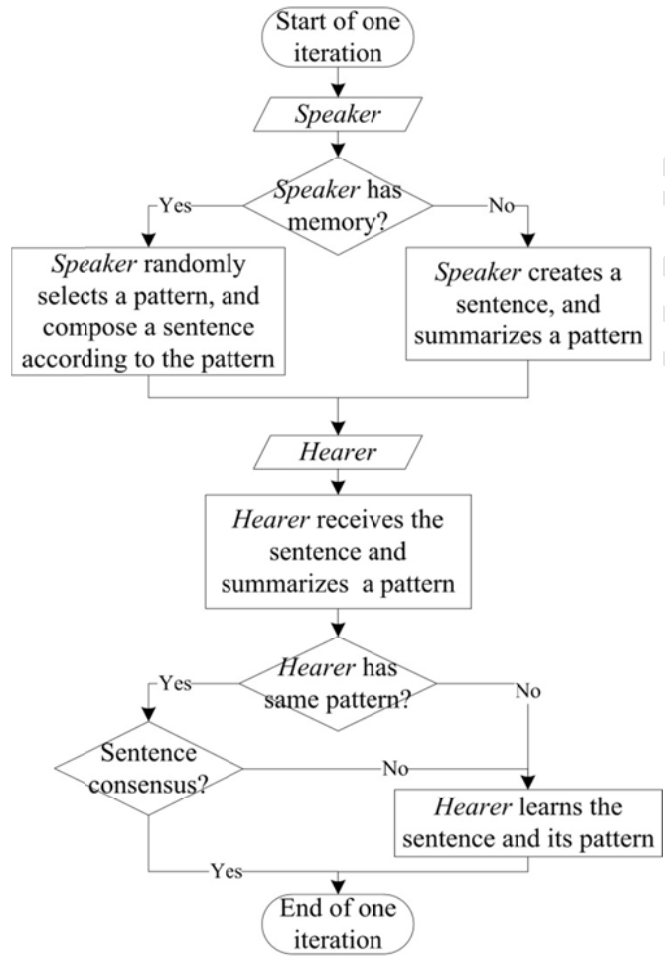

**Figure 2** Flowchart of one iteration in the MWNG. Iteration starts with a randomly selected speaker-hearer pair from the population. The speaker utters a sentence taken from his memory, if any; otherwise, he generates a new one. The hearer receives the sentence and then checks if he has the same pattern stored in his memory. Only if the hearer has the same pattern in the memory, he will verify the sentence word by word; otherwise, he learns the pattern and/or (some) words of the received sentence. When the hearer has the same sentence pattern and all words as the speaker uttered, they reach a local consensus, so both speaker and hearer will keep the consented sentence (including the consented pattern and all the consented words) in their memories, while dropping all other patterns and words therein.

P1: Subject + Verb (*S.* + *V.*)
    -- *Cat runs.*
P2: Subject + Verb + Object (*S.* + *V.* + *O.*)
    -- *Boys play football.*
P3: Subject + Verb + Complement (*S.* + *V.* + *C.*)
    -- *Flowers are beautiful.*
P4: Subject + Verb + indirect Object + direct Object (*S.* + *V.* + *iO.* + *dO.*)
    -- *She teaches students English.*
P5: Subject + Verb + Object + Complement (*S.* + *V.* + *O.* + *C.*)
    -- *People elect him presisent.*

**Figure 3** Five simple and conventional English language patterns. They are denoted by P1 to P5. Abbreviations are given inside the following parentheses and an example sentence is given to each pattern therein. Note that in P4, there are two components belonging to the category '*object*'. In the simulation, we implement '*object*' into two subsets, one for '*indirect*

*object*' and the other for '*direct object*'. When communicating with pattern P4, two '*object*' subsets are treated independently. When communicating with pattern P2 or P5, the two subsets are combined as one set.

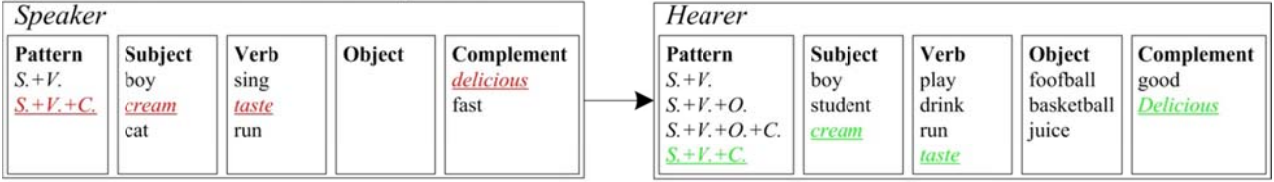
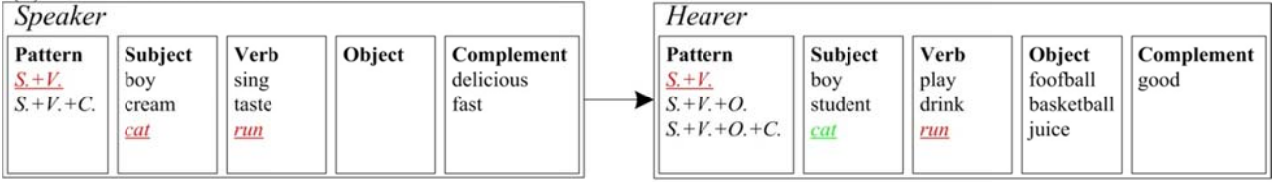
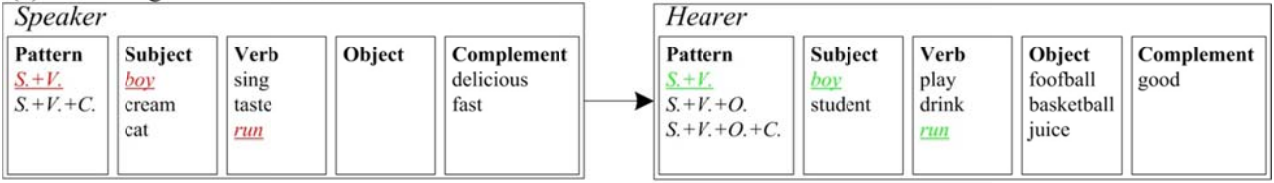
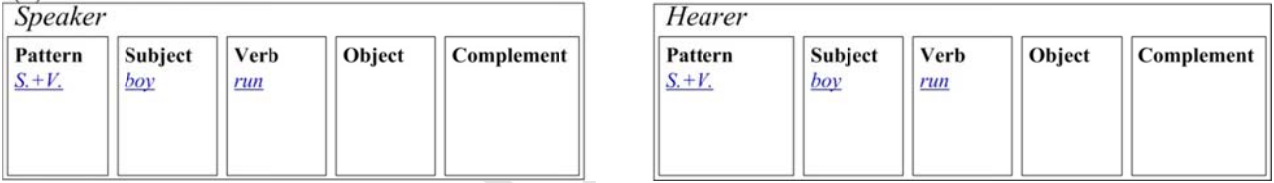

**Figure 4** An example of local pair-wise communication. (a) Speaker randomly chooses the pattern '*S.+V.+C.*'(P3) and three words to utter a sentence '*Cream tastes delicious*'. Since Hearer has no this same pattern P3, he learns the pattern, as well as every word in the sentence. (b) Although Hearer has the same pattern P1 as Speaker uttered, he does not have an identical sentence; so he learns the words in the sentence. Note that since Hearer has already had the word '*cat*' in his memory, he would neither learn it again nor have consent to this single word. (c) Hearer has the same pattern P1, as well as all the words in the sentence. (d) The state when Speaker and Hearer reach pair-wise local consensus: both Speaker and Hearer have only pattern P1 and the agreed sentence '*boy run(s)*' in their memories.

**Conventional sentence patterns.** We first choose five simple conventional English language patterns to study the sentence propagation in MWNG. The vocabulary is classified into four categories: '*subject*', '*verb*', '*object*', and '*complement*', as shown in Figure 3.

The real-life situation is complicated. A sentence is usually with a context supported by such as conversation and background information, etc. The context helps disambiguate the conversation, so that even if the speaker and hearer can at least reach a partial-consensus state. However, in NG simulations, we do not offer such useful contexts to the agents. Further, an MWNG model with a partial-consensus rule is essentially equivalent to a parallel combination of multiple SWNGs. As such, the local partial-consensus is ruled out. In MWNG, it requires the hearer to achieve consensus with the speaker with respect to both every single word in the sentence as well as the sentence pattern, so as to guarantee the correct meaning of the saying; otherwise, they fail reaching consensus in the interaction.

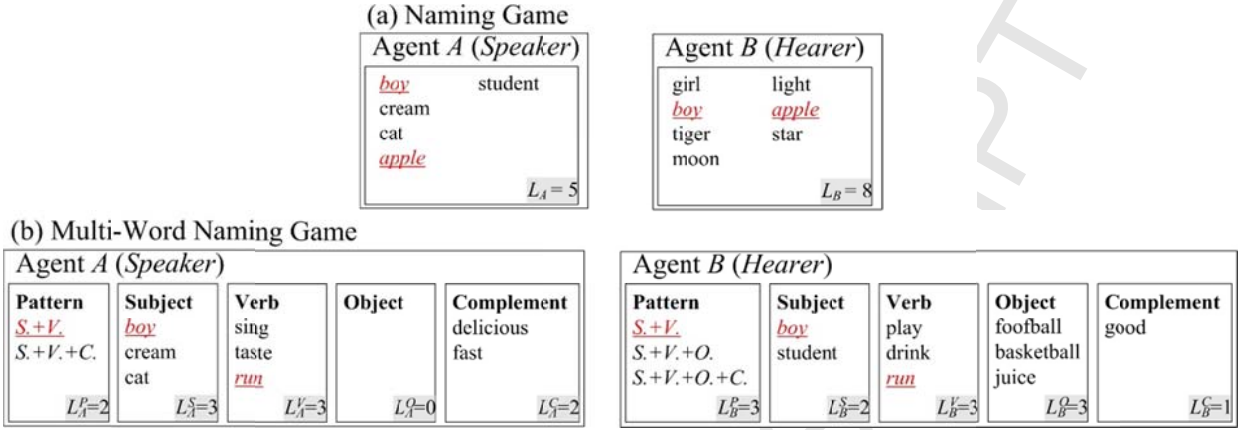

**Figure 5** An example to illustrate the probability of consensus within one communication in (a) the minimal NG and (b) MWNG. The memory length is shown at the lower-right corner of each agent's memory box.

**Local consensus.** Figure 4 shows some examples of pair-wise communications: Figure 4 (a) shows the hearer learns a new pattern from the speaker. Figure 4 (b) shows when the hearer does not hold every single word of the transmitted sentence in his memory. Figure 4 (c) shows local consensus and Figure 4 (d) is the result of local consensus.

Suppose that, at a time step of the minimal NG, two connected agents $A$ and $B$ have been picked up as speaker and hearer, respectively. Agent $A$ has $L_A$ words in memory while agent $B$ has $L_B$ words. The number of common words that both $A$ and $B$ have is $I_{AB}$. The probability of consensus within one communication is $P_{NG} = \frac{I_{AB}}{L_A}$. As for MWNG, an agent has several separated parts of memory to store patterns and words of different categories. Denote the memory lengths for patterns by $L_A^P$ (for agent $A$) and $L_B^P$ (for agent $B$). All the words $W$ are evenly divided into $M$ categories, i.e., $W = \{W_1, W_2, ..., W_M\}$, thus, the numbers of words remembered by agents $A$ and $B$ are denoted by $L_A^{W_i}$ and $L_B^{W_i}$ ($i = 1,2, ..., M$), respectively. The number of common patterns that both $A$ and $B$ have is denoted by $I_{AB}^P$, and the numbers of common words are denoted by $I_{AB}^{W_i}$ ($i = 1,2, ..., M$). The probability of consensus within one communication is

$$P_{MWNG} = \rho \frac{I_{AB}^P}{L_A^P} \qquad (1)$$

where $\rho = \prod_{j \in R} \frac{I_{AB}^{W_j}}{L_A^{W_j}}$, $R$ ($R \subset W$) represents the component word categories of the picked pattern.

An example is given in Figure 5. The number of remembered words (or patterns) can be seen at the lower-right corner of each agent's memory box. Figure 5 (a) shows the minimal NG case, where agents $A$ and $B$ have two words in common ($I_{AB} = 2$), thus, the probability of consensus in one communication is $P_{NG} = \frac{I_{AB}}{L_A} = 0.4$. In Figure 5 (b), two agents have one pattern in common ($I_{AB}^P = 1$). There are two word categories employed in this pattern, namely $R = \{S, V\}$. $R \subset W = \{S, V, O, C\}$. Both agents have one common word in '*subject*' and '*verb*', respectively. Thus, $I_{AB}^S = 1$ and $I_{AB}^V = 1$, then $\rho = \prod_{j \in R} \frac{I_{AB}^{W_j}}{L_A^{W_j}} = \frac{I_{AB}^S}{L_A^S} \times \frac{I_{AB}^V}{L_A^V} = \frac{1}{9}$, thus $P_{MWNG} = \rho \frac{I_{AB}^P}{L_A^P} = \frac{1}{18}$.

The probability of consensus of a sentence in MWNG equals to the probability of consensus of a pattern multiplies all the probabilities of consensus of the related words. The probability of consensus within one communication in MWNG is lower than that in the minimal NG, but nothing prevents consensus.

**Table 1** Network settings in simulations. A total of 12 random-graph (RG), small-world (SW) and scale-free (SF) networks of 500 nodes each are employed for simulation. The networks are randomly generated and the basic properties including average node degree, average path length, and average clustering coefficient are obtained by averaging over 30 independent runs. The network settings of 1000 and 1200 nodes, as well as the corresponding simulation results, are given in the SI [23].

| Notation | Network type | Number of nodes | Average degree | Average path length | Average clustering coefficient |
|---|---|---|---|---|---|
| *RG*/0.03 | *Random-graph* network with $P = 0.03$ | 500 | 14.98 | 2.5956 | 0.0302 |
| *RG*/0.05 | *Random-graph* network with $P = 0.05$ | 500 | 24.97 | 2.2228 | 0.0500 |
| *RG*/0.1 | *Random-graph* network with $P = 0.1$ | 500 | 49.98 | 1.9057 | 0.1002 |
| *SW*/50/0.1 | *Small-world* network with $K = 50$ and $RP = 0.1$ | 500 | 100 | 1.8049 | 0.5676 |
| *SW*/50/0.2 | *Small-world* network with $K = 50$ and $RP = 0.2$ | 500 | 100 | 1.7997 | 0.4382 |
| *SW*/50/0.3 | *Small-world* network with $K = 50$ and $RP = 0.3$ | 500 | 100 | 1.7996 | 0.3457 |
| *SW*/60/0.1 | *Small-world* network with $K = 60$ and $RP = 0.1$ | 500 | 120 | 1.7599 | 0.5725 |
| *SW*/60/0.2 | *Small-world* network with $K = 60$ and $RP = 0.2$ | 500 | 120 | 1.7595 | 0.4521 |
| *SW*/60/0.3 | *Small-world* network with $K = 60$ and $RP = 0.3$ | 500 | 120 | 1.7595 | 0.3672 |
| *SF*/25 | *Scale-free* with 26 initial nodes and 25 new edges added at each step | 500 | 48.64 | 1.9272 | 0.1972 |
| *SF*/50 | *Scale-free* with 51 initial nodes and 50 new edges added at each step | 500 | 94.81 | 1.8102 | 0.3088 |
| *SF*/75 | *Scale-free* with 76 initial nodes and 75 new edges added at each step | 500 | 138.47 | 1.7228 | 0.3983 |

## 3. Results

**Simulation setup.** Large-scale numerical simulations are carried out on three typical network topologies, namely, random-graph (RG) [6,20], small-world (SW) [8,9,21] and scale-free (SF) [6,7] networks. The performances of emergence, propagation and consensus of sentences and their patterns are examined. Simulation data reflecting different aspects of NG are collected with comparisons. The agents store nothing in their memories initially, and the memory size of each agent is large enough or infinite [13,22]. Totally 5 conventional patterns in English language are used (shown in Figure 3) to form various sentences. A total of 12 settings of the communication networks are investigated, each with 500 nodes (agents). The settings and basic properties of the networks are summarized in Table 1. To further examine the scaling property of the population size, the study on these 12 networks of 1000 and 1200 nodes are presented in the Supplementary Information (SI) [23]. In addition, 5 sets of man-designed test patterns are used to examine the convergence of the sentence patterns.

**Convergence process of conventional English language patterns.** All the subplots in Figures 6, 7, 8 and 9 are in the same coordinate, so that the curves of different subplots can be compared easily and directly.

Figure 6 shows the convergence process with the number of total words in the population. It can be seen that the convergence process has a first-increase-then-decrease curve. For each category of words, it is similar to the SWNG, but with slight oscillations between the saturation and converging phrases. For the minimal NG on random-graph and small-world networks, the maximum number of total names is given by $N_{name}^{max} = \frac{N}{2}\left(1 + \frac{\langle k \rangle}{2}\right)$, where $N$ is the population size and $\langle k \rangle$ is the average degree of the underlying network [2]. For example, for RG/0.03 ($N = 500$, $\langle k \rangle = 14.98$ as shown in Table 1, the maximum number of total names is $N_{name}^{max} = 2123$. Comparing with Figure 6 (A), it is obvious that the maximum number of total names of each category is more than two thousands, and that of '*subject*' and '*verb*' is up to some six thousands. The calculation of the maximum number of total names in a scale-free network can also be found in [2]. Noticeably, the sentence patterns and multiple words here make the communications more complicated and require the agents to store more names throughout the process. During the saturation-convergence transition phase, local consensus suffers more from disturbing when the atomic names are divided into several categories. This complication produces the oscillatory behavior making local consensus as well as global consensus difficult to take place. The success rate curve for each network is plotted as background in Figure 6. One can see that, during the iteration steps between $10^4$ and $10^5$, as the success rate slightly increases, the number of total words drastically decreases. This is because, on the average, each agent has accumulated many names in his memory, and local consensus may require clearing up tens of names therein. When the tendency of global consensus is prominent, the success rate curve raises and the number of total names drops both drastically and smoothly.

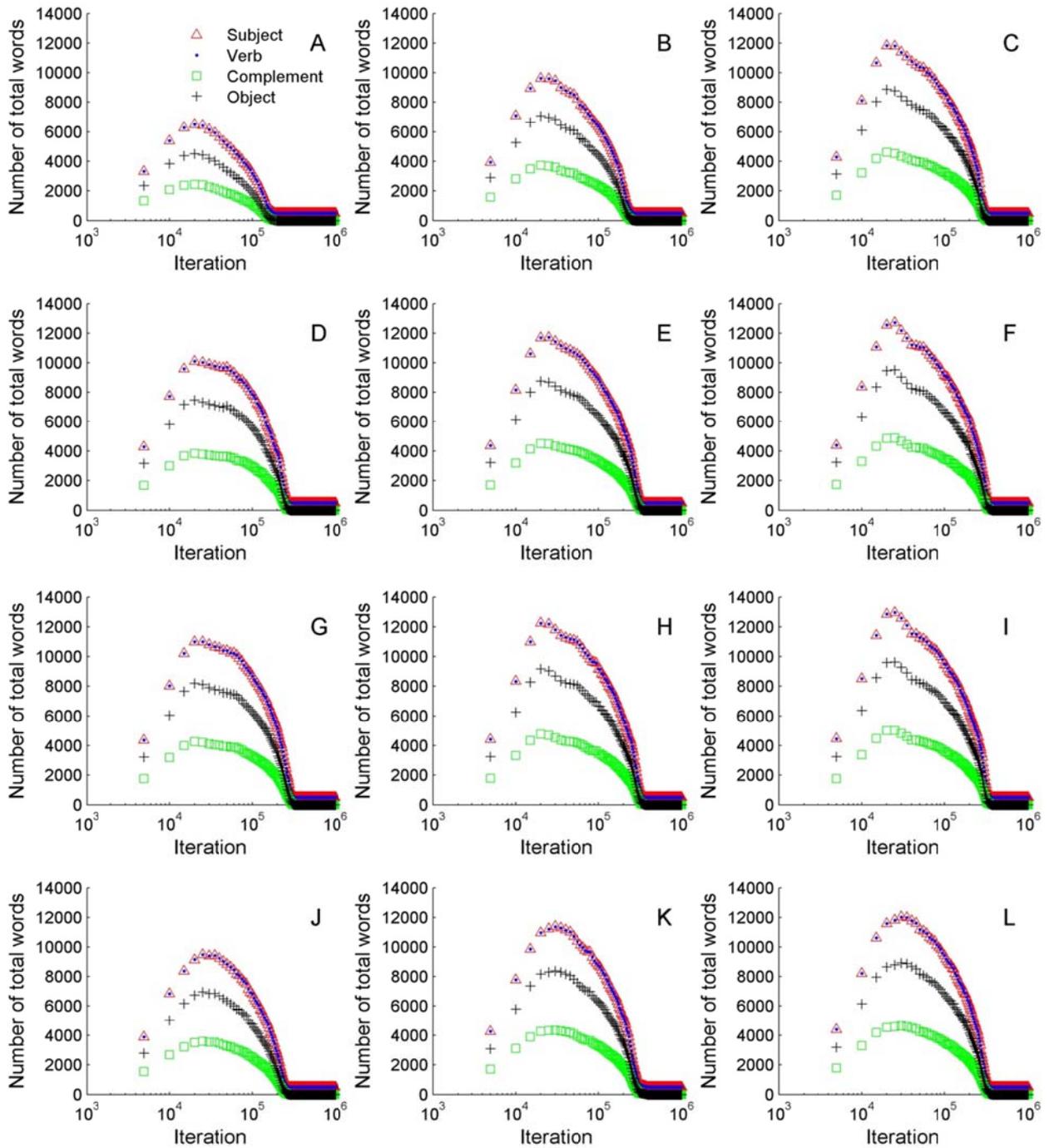

**Figure 6** Convergence curves in terms of the number of total words vs. iterations, accompanied with the success rate, used as reference: (A) RG/0.03; (B) RG/0.05; (C) RG/0.1; (D) SW/50/0.1; (E) SW/50/0.2; (F) SW/50/0.3; (G) SW/60/0.1; (H) SW/60/0.2; (I) SW/60/0.3; (J) SF/25; (K) SF/50; (L) SF/75. In each subfigure, the converging process is plotted as 4 curves, representing 4 categories of words, respectively. The process of success rate is included as background for reference. The comparison between the success rate curves is shown in Figure 9. Since in the tested conventional English sentence (sub-)pattern, '*subject*' and '*verb*' always appear together, the convergence curves of them are consistent with each other. For each of the 4 types of underlying networks (on each row), the parameters of (re-)connection probability as well as average node degrees increase from left to right, and the peaks of the curves become higher from left to right. The convergence time becomes longer as the network parameter values increase. Note that the numbers of total words for '*complement*' and/or '*object*' are zero when the population finally reaches a consent rule without these types of words, while the numbers of both '*subject*' and '*verb*' eventually reach the population size, 500.

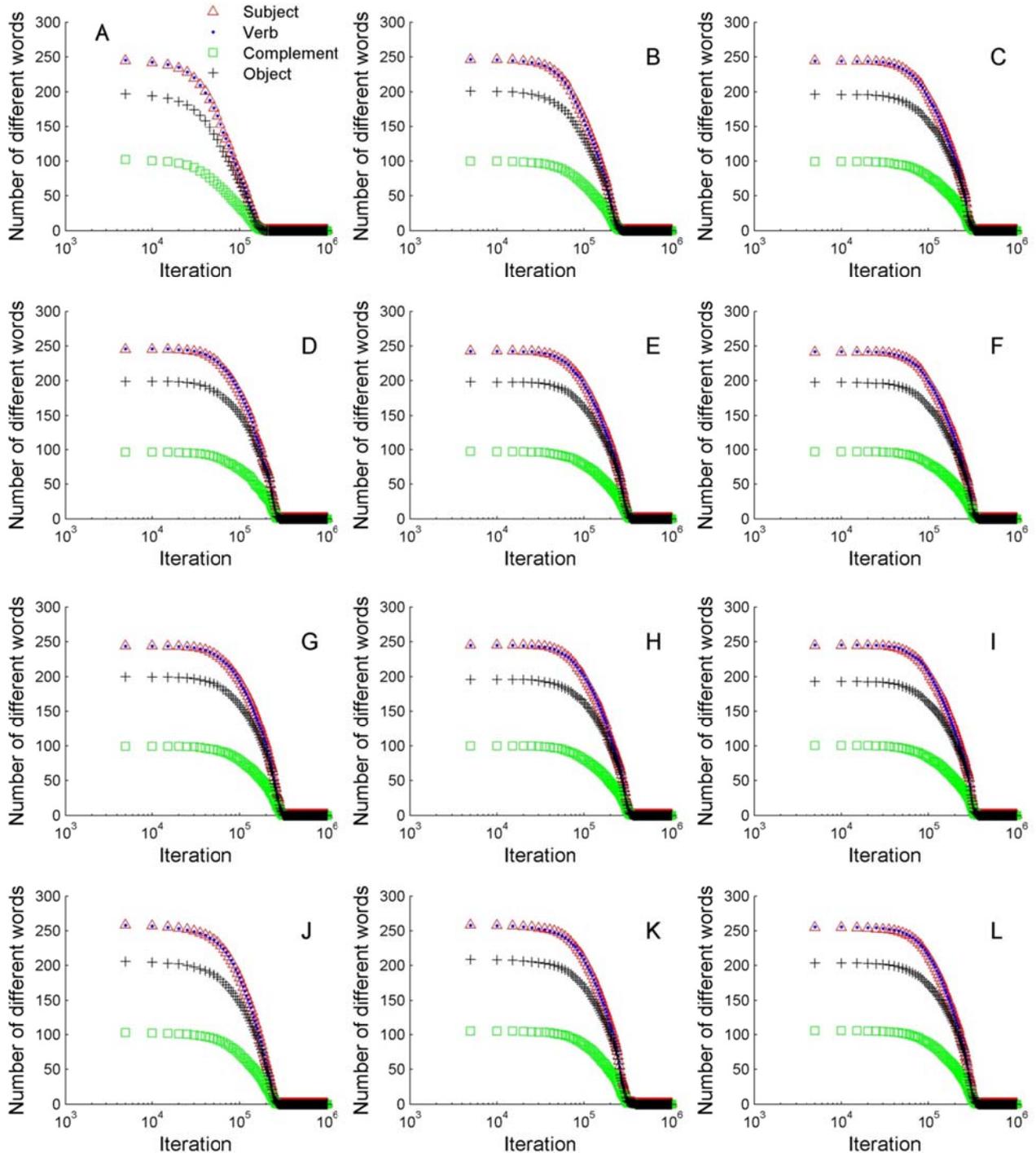

**Figure 7** Convergence curves in terms of the number of different words vs. iterations: (A) RG/0.03; (B) RG/0.05; (C) RG/0.1; (D) SW/50/0.1; (E) SW/50/0.2; (F) SW/50/0.3; (G) SW/60/0.1; (H) SW/60/0.2; (I) SW/60/0.3; (J) SF/25; (K) SF/50; (L) SF/75. Differing from the curves of the number of total words, when the network parameters are changed, the shapes of the curves are nearly unchanged but slightly shifted. Moreover, since all the curves are plotted in identical coordinates, they can be compared vertically. There is hardly any difference in curves between different network types.

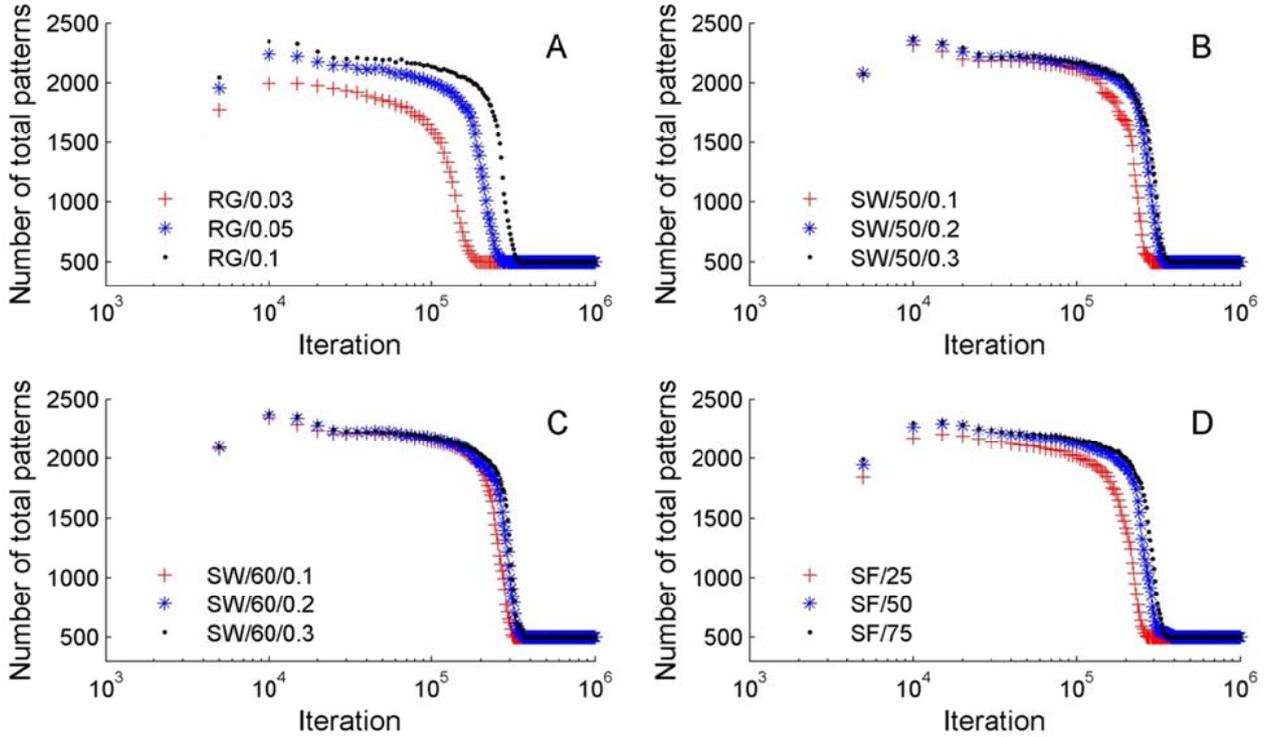

**Figure 8** Convergence curves in terms of the number of total patterns vs. iterations: (A) Random-graph networks; (B) and (C) Small-world networks; (D) Scale-free networks. When the (re-)connection probability is small, the peak of a curve is lower and the convergence takes place earlier. Except for the curve of RG/0.03, shown in (A), the peaks of other curves are higher than 2000, but (slightly) lower than 2500 (as shown in Figure 10, which is about 2400). This means that there is a time period where the agents have learned more than 4 patterns on average. Many agents have learned all 5 patterns.

Figure 7 shows the convergence curves in terms of the number of different words, where all 12 subplots have very similar shapes to each other. The peaks of '*subject*' and '*verb*' are both around 250, the peak of '*complement*' is around 100, and the peak of '*object*' is around 200. This matches the fact shown in Figure 3 that, '*subject*' and '*verb*' appear in each of the five patterns, '*complement*' appears twice in all five patterns. Though '*object*' appears in three patterns, it appears four times in all five patterns. Therefore, the maximum number of different '*complement*' words is 40% of the maximum number of different '*subject*' or '*verb*' words, whereas for the maximum number of different '*object*' words, it is 80% of the maximum number of different '*subject*' or '*verb*' words.

Figure 8 shows the curves of the number of total patterns. The curves show that, during a time period, the agents have learned more than four patterns on average, which suggests $\frac{l_{AB}^P}{L_A^P} > 0.8$ in Equation (1).

Figure 9 shows the success rate curves. The success rate reflects the ratio of local consensus in the last ten time steps [11,13]. Because the definition of local consensus avoids partial-consensus, as indicated above, the consensus process is drastic rather than gradual.

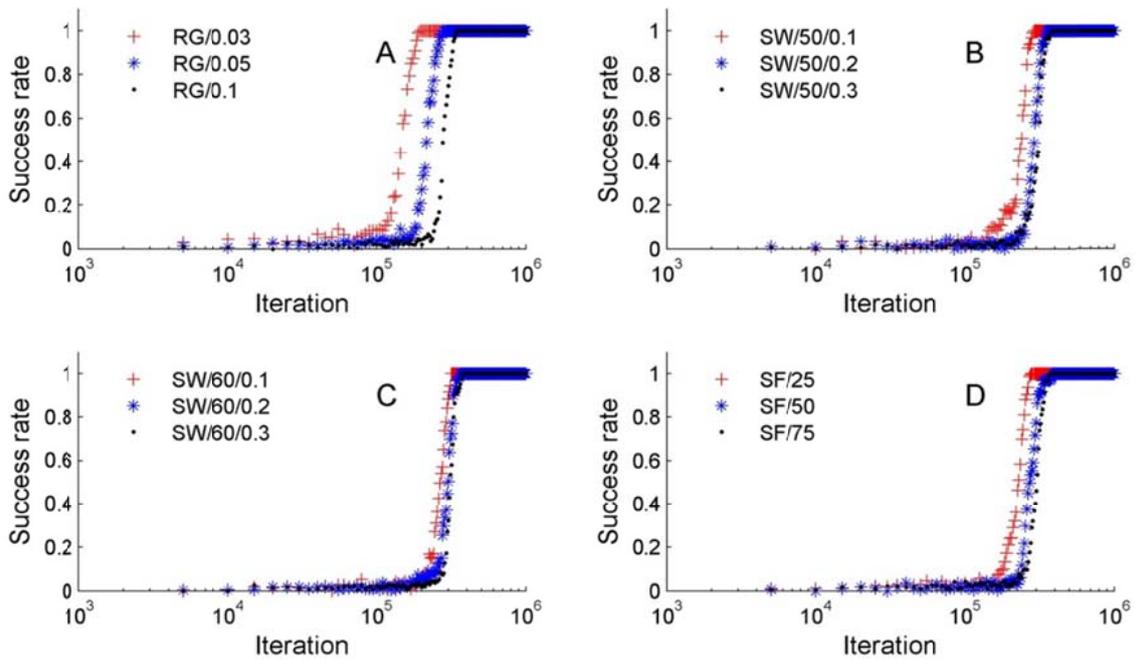

**Figure 9**   Curves of the success rate: (A) Random-graph networks; (B) and (C) Small-world networks; (D) Scale-free networks. The success rate curves of MWNG are simple as compared with the oscillatory success rate curves of small-world networks in SWNG [13]. Before global converge takes place, the success rate stays below 0.2; then, in the converging phase, the success rate increases dramatically to reach 1.0. Note that all these 12 success rate curves are also plotted in Figure 6 for reference.

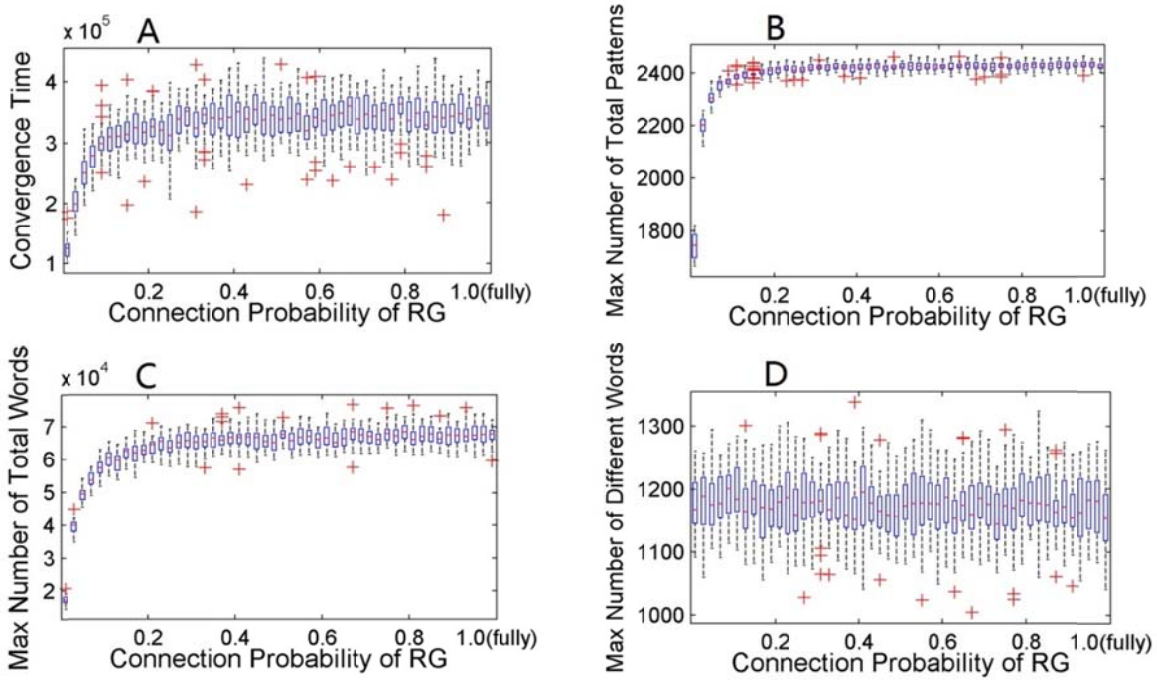

**Figure 10**   Study on the random-graph of 500 nodes with the connection probability varying from 0.02 to 1.0 (incremental step size 0.02): (A) Convergence time; (B) Maximum number of total patterns; (C) Maximum number of total words; (D) Maximum number of different words. When the connection probability is between 0.02 and nearly 0.4, the convergence time and maximum numbers of total patterns and total words all increase as the connection probability increases. When the connection probability is greater than 0.4, these three indexes become plateaued. In contrast, the maximum number of different works is not affected by the change of the connection probability. All the curves are averaged over 30 independent runs.

Figure 10 gives box plots of 4 indexes for convergence time, maximum number of total patterns, maximum number of total words, and maximum number of different words, against the change of the connection probability, in a 500-node random-graph communication network. In each box plot, the blue box denotes that the central 50% data lies within this section; the red bar is the median value of all 30 independent datasets; the upper and lower black bars are the greatest and the least values, excluding outliers which are indicated by red pluses. Figure 10 (A) shows that, in the MWNG model, the convergence time is non-decreasing as the connection probability (as well the average node degree) increases. This behavior is quite different from the atomic NG, as reported in [2], where the convergence time decreases monotonously as the average node degree increases. A greater value of the average node degree introduces more information input to every agent on the average; as a result, the agents in a better connected network will accumulate more names than those connected on a network with lower average node degree.

However, in a SWNG model, the number of accumulated names in the agents' memories will directly influence the achieving probability of local consensus. The reason is as follows: 1) the number of different words is limited and not affected by the average node degree. Only when an agent has nothing in his memory will he invent a name. Figure 10 (D) supports this viewpoint, albeit empirically. 2) The more names the agents have accumulated, the more common names they will have, so that a higher probability of local consensus will be gained. As a result, the convergence time decreases monotonously as the average node degree increases.

In MWNG, accumulating more names will not directly influence the probability of reaching local consensus. Its impact on the probability of reaching local consensus is even lower as compared to the SWNG due to the conflict in pairing words from different categories, which does not exist in the latter. As can be seen from the simulation results to be reported in the next subsection, the more components a pattern has, the more difficult achieving consensus on this pattern will be. As a result, a better connection in the underlying communication network does not facilitate the convergence speed of MWNG in general.

**Convergence process of man-designed patterns.** In the following simulations, five man-designed language pattern sets are considered, with each set includes three to six patterns. The conventional English language patterns are natural and efficient in real-life communications, but not so in experimental studies, for instance the categories '*subject*' and '*verb*' appear in the beginning of all five patterns in the same ordering. In all the above-reported experimental results, the population always converges to this simplest '*subject+verb*' (P1) pattern. The man-designed pattern sets are designed for further study on convergence.

In the literature, as said by William Shakespeare, "*brevity is the soul of wit*". Correspondingly in scientific research, as reported in [24], "*papers with shorter titles receive more citations per paper*". To a certain extent, (recognition of) '*wit*' and '*citation*' can be also considered as *one-sided consensus*, since neither (recognition of) '*wit*' nor '*citation*' is the result of mutual interactions as the local or global *consensus* in NG.

Differing from the literature and also scientific research, for which the reasons for short expressions are still unclear [24], the reason for MWNG to converge to shorter sentence patterns is clear, and indeed quite simple: a shorter pattern has more opportunities to reach consensus than those longer ones. For example, in one time step, assuming that both the speaker and the hearer store some identical $N_s$ '*subjects*', $N_v$ '*verbs*', $N_o$ '*objects*' and $N_c$ '*complements*' in their memories, and both have learned all five patterns (P1-P5). Let the speaker and the hearer have identical memory size, which is the optimal situation for one-step consensus. If the speaker utters a P1 sentence, the probability of reaching local consensus is $\frac{1}{N_s \cdot N_v}$, while the probability of uttering a P5 sentence is $\frac{1}{N_s \cdot N_v \cdot N_o \cdot N_c}$, where obviously $\frac{1}{N_s \cdot N_v} > \frac{1}{N_s \cdot N_v \cdot N_o \cdot N_c}$. If no assumption on identical memory is made, for example, when an agent learned '*boys play football*', he also has consensus with '*boys play*', but not vice versa. As a result, shorter pattern has greater probability to reach consensus and to survive eventually.

Note that this phenomenon in MWNG cannot directly explain why papers with shorter titles receive more citations. This is because, in NG, the meaning of a sentence is not as clear and precise as a paper title, and furthermore, '*citation*' is different to infer from '*consensus*' in NG. In NG, the population tries to name one single object, randomly and uniformly, but the paper titles bare significant information for expression. Nevertheless, MWNG can provide some hints for the phenomenon that, if the amount of information in a sentence is uniform to describe an object, the agents are prone to accept shorter ones, although five patterns are randomly and uniformly chosen by them.

The results of a statistical study on the eventually converged man-designed patterns are shown in Table 2. The five man-designed pattern sets are denoted as A, B, C, D and E in Figure 11. In each pattern set, define several test patterns (TP), and each pattern composes of several test categories (Tc). The modifier *test* (T) is used to distinguish the conventional English language patterns (P1~P5). The test pattern sets are man-designed, used for testing the eventually converged pattern distributions. Sets A and B are uniformly distributed. The difference between Set C and Set D is that, in Set D, TP1 is a component ('*Tc1+Tc2*') of TP4, which does not exist in Set C. In Set E, the longest pattern shares no common parts with other (shorter) patterns.

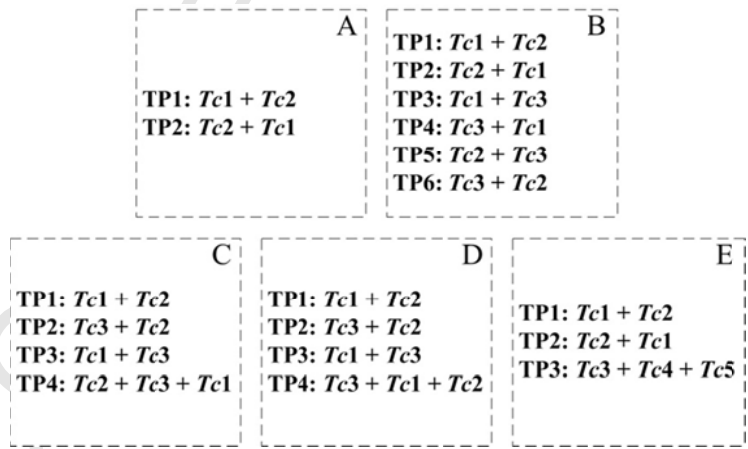

**Figure 11** The five man-designed test categories of words (Tc1-Tc5) and five test pattern sets are designed: (A) Two uniform patterns with two word categories; (B) Six uniform patterns with three word categories; (C) and (D) Two sets of patterns of different lengths; (E) Three patterns with five word categories. The pattern sets in (A) and (B) are uniformly distributed. In (C), none of the shorter patterns is a sub-sequence of the longer pattern TP4, while in (D), TP1 is a sub-sequence of TP4. The pattern set in (E) includes three patterns, where TP3 shares no common word categories with other patterns.

Simulations on 500 and 1000 population sizes are implemented and the results are summarized in Table 2. It can be observed: 1) From the results of Sets A and B, the eventually converged patterns are uniformly distributed, when the patterns are of equal length and the categories are uniformly distributed. 2) From the results of Sets C and D, the longer pattern (TP4) has no chance to be converged to. TP4 of Set D contains a sub-pattern "$Tc1+Tc2$", which is TP1 of Set D, while TP4 of Set C does not contain such a sub-pattern. 3) From the results of Set E, when a longer pattern (TP3) shares no common word categories with shorter patterns, it will be converged to, but with a small probability.

**Table 2** The number of eventually converged patterns in 5 test sets (these man-designed pattern sets are defined in Figure 11). There are 12 networks simulated over 30 independent runs, thus there are 360 trials in total. Each integer represents the number of trials which led the population to converge to that pattern, with its proportion indicated in the parentheses.

| Number of nodes | Test pattern set | TP 1 | TP 2 | TP 3 | TP 4 | TP 5 | TP 6 |
|---|---|---|---|---|---|---|---|
| 500 | A | 183 (0.51) | 177 (0.49) | / | / | / | / |
|  | B | 63 (0.17) | 69 (0.19) | 49 (0.14) | 61 (0.17) | 65 (0.18) | 53 (0.15) |
|  | C | 119 (0.33) | 139 (0.39) | 102 (0.28) | 0 (0.00) | / | / |
|  | D | 125 (0.35) | 123 (0.34) | 112 (0.31) | 0 (0.00) | / | / |
|  | E | 174 (0.48) | 167 (0.47) | 19 (0.05) | / | / | / |
| 1000 | A | 171 (0.48) | 189 (0.52) | / | / | / | / |
|  | B | 61 (0.17) | 66 (0.18) | 64 (0.18) | 57 (0.16) | 53 (0.15) | 59 (0.16) |
|  | C | 106 (0.29) | 132 (0.37) | 122 (0.34) | 0 (0.00) | / | / |
|  | D | 117 (0.33) | 124 (0.34) | 119 (0.33) | 0 (0.00) | / | / |
|  | E | 156 (0.43) | 189 (0.53) | 15 (0.04) | / | / | / |

## 4. Discussion

In this paper, we proposed a multi-word naming game (MWNG) and studied it by means of extensive and comprehensive computer simulations. MWNG is a new model simulating the situation where a population of social agents tries to invent, propagate and learn to describe a single object (opinion or event) by a sentence of several words in a language pattern. We studied MWNG on five conventional English language patterns and five man-designed test pattern sets. The simulation results show that: 1) the new sentence sharing model is an extension of the classical lexicon sharing model, in which their processes and features are basically similar; 2) the propagating, learning and converging processes are more complicated than that in the conventional NG, since larger memory size and longer convergence time are needed in MWNG; 3) the convergence time is non-decreasing as the network becomes better connected, while greater value of average node degree reduces the convergence time in the single-word naming game (SWNG); 4) the agents are prone to accept short sentence patterns, consistent with many known linguistic phenomena in the real world. These new findings may help to enhance our understanding of the human language emergence and evolution from a network science perspective.

**Acknowledgement**

This research was supported by the Hong Kong Research Grants Council under the GRF Grant CityU11208515, and the National Natural Science Foundation of China under Grant 61473321.

**Highlights**

- A new model "multi-word naming game (MWNG)" is proposed.
- MWNG extends single-word propagation to sentence propagation.
- Simulations are implemented on random-graph, small-world and scale-free networks.